# The great AI witch hunt: Reviewers' perception and (Mis)conception of generative AI in research writing

Hilda Hadan, Derrick M. Wang, Reza Hadi Mogavi, Joseph Tu, Leah Zhang-Kennedy, Lennart E. Nacke *

*Stratford School of Interaction Design and Business, University of Waterloo, Waterloo, Ontario, N2L 3G1, Canada*

ABSTRACT

Generative AI (GenAI) use in research writing is growing fast. However, it is unclear how peer reviewers recognize or misjudge AI-augmented manuscripts. To investigate the impact of AI-augmented writing on peer reviews, we conducted a snippet-based online survey with 17 peer reviewers from top-tier HCI conferences. Our findings indicate that while AI-augmented writing improves readability, language diversity, and informativeness, it often lacks research details and reflective insights from authors. Reviewers consistently struggled to distinguish between human and AI-augmented writing but their judgements remained consistent. They noted the loss of a "human touch" and subjective expressions in AI-augmented writing. Based on our findings, we advocate for reviewer guidelines that promote impartial evaluations of submissions, regardless of any personal biases towards GenAI. The quality of the research itself should remain a priority in reviews, regardless of any preconceived notions about the tools used to create it. We emphasize that researchers must maintain their authorship and control over the writing process, even when using GenAI's assistance.

## 1. Introduction

The emergence of generative artificial intelligence (GenAI) tools such as ChatGPT[1] and Gemini[2] have sparked a wave of excitement in academia and industry. Since the release of ChatGPT in November 2022 (OpenAI, 2022), GenAI has become increasingly popular in assisting people with written, auditory, and visual tasks (Hadi Mogavi et al., 2024; Mohamed et al., 2024; Tu et al., 2024). In research, GenAI offers a new approach to manuscript writing, as it can handle tasks ranging from text improvement suggestions to speech-to-text translation and even crafting initial drafts (Lin et al., 2024; Mohamed et al., 2024). Its ability to understand context and generate human-like and grammatically accurate responses fosters innovative brainstorming and enhances the quality and readability of research publications (Babl & Babl, 2023). However, along with GenAI's potential to augment research activities, concerns about transparency, academic integrity, and the urgency of maintaining the credibility of research work have emerged (Committee on Publication Ethics [COPE], 2023b; Liu et al., 2020; Cruz Rivera et al., 2020; Tu et al., 2024).

Despite the growing interest in using GenAI for manuscript writing and research activities (Mohamed et al., 2024; Phillips, Trick, Nacke, & Mandryk, 2023), many researchers hesitate to acknowledge its use in their papers. This is illustrated by several instances where research publications with undisclosed GenAI use were identified by readers (e.g., LinkedIn.com, 2024; Reddit.com, 2024; Retraction Watch; Twitter.com, 2023). Studies have identified the phenomenon of AI aversion, where AI-generated content, even if factual, is often perceived as inaccurate and misleading (Burton, Stein, & Jensen, 2020; Longoni, Fradkin, Cian, & Pennycook, 2022) and disclosing its use can negatively impact readers' satisfaction and perception of the authors' qualifications and effort (Irene, 2024). Therefore, researchers' hesitancy is partly due to their fear that acknowledging GenAI use might damage reviewers' perceptions. However, given the widespread adoption of GenAI, researchers' undisclosed GenAI use will harm the transparency, credibility, and integrity in research knowledge mobilization in the long-term.

Our research investigates perceptions of academia and industry professionals experienced in peer-reviewing manuscripts for top-tier human-computer interaction (HCI) conferences. Through understanding reviewers' perceptions and clarifying their possible misconceptions, we seek to reduce researchers' concerns about disclosing GenAI use. Our findings will shed light on the impacts of using GenAI as writing assistance for both reviewers and researchers, and foster a transparent and credible research environment. Specifically, we answer four Research Questions (RQs).

* Corresponding author. Stratford School of Interaction Design and Business, University of Waterloo, 125 St Patrick St, Stratford, ON, N5A 0C1, Canada.
*E-mail address:* lennart.nacke@acm.org (L.E. Nacke).

[1] ChatGPT. https://chat.openai.com/.
[2] Gemini (formerly Google Brad). https://gemini.google.com/.

**RQ1**. *How much are reviewers aware of the use of AI in the context of research manuscripts?*

A recent study has identified concerns among researchers about their writing being indistinguishable from AI-generated text, especially for those trained in formal writing structures (Tu et al., 2024). In fact, non-native English writing samples are more likely to be misclassified as AI-generated (Liang, Yuksekgonul, Mao, Wu, & Zou, 2023), and human cannot differentiate between AI- and human-written content (Gao et al., 2023). Therefore, false positives might occur among peer reviewers' assessment of manuscripts. Our RQ1 aims to validate this hypothesis by examining reviewers' awareness across various levels of AI involvement in research writing.

**RQ2**. *How much is reviewers' judgement on research and manuscript quality influenced by the use of AI in its writing?*

The phenomenon of AI aversion (Burton et al., 2020; Longoni et al., 2022) further raises the issue that reviewers might be biased in their assessment of the quality and credibility of the research presented in submissions. Our RQ2 aims to explore this issue by examining how snippets with various levels of AI involvement in writing influence reviewers' judgments.

**RQ3**. *To what extent do reviewers' peer-review experience, disciplinary expertise, and AI familiarity influence their perception and judgement?*

Literature suggests that people's familiarity with algorithms and expertise in relevant fields shape their perceptions (Berkeley, Simmons, & Massey, 2015; Graefe, Haim, Haarmann, & Brosius, 2018; Logg, Minson, & Moore, 2019). Therefore, reviewers' peer-review experience, disciplinary expertise, and familiarity with GenAI may also shape their perceptions. Our RQ3 aims to investigate how these factors impact reviewers' perceptions and judgments.

**RQ4**. *What aspects of research writing impact reviewers' perception and judgement?*

Prior research indicates that GPT detectors often misclassify content with limited linguistic proficiency as AI-generated (Liang et al., 2023), and that human-authored articles are generally seen as more pleasant to read and less boring (Christer, 2017). Our RQ4 seeks to identify specific manuscript's elements that shape reviewers' perceptions. Through identifying these elements, we aim to uncover the rationale behind reviewers' judgments and misconceptions about GenAI in manuscript writing.

We investigated peer-reviewer perception through an online survey. To the best of our knowledge, our study is the first to empirically examine how peer-reviewers from top-tier HCI conferences perceive AI-augmented academic writing across three types of text: original human-written, AI-paraphrased, and AI-generated snippets. Our approach for assessing peer-reviewer perceptions of AI-augmented writing can be adapted for use in other academic fields than HCI. While our research is focused on HCI, it has broader implications for academic publishing across disciplines. We offer insights into the relationships of GenAI, authorship, and peer review. Our research makes four additional contributions to research on GenAI-augmented manuscript writing and its regulation. *First*, we show that all peer-reviewers struggled to distinguish between AI-processed and human-written snippet. All reviewers perceived AI-paraphrased snippets as more honest. Reviewers with more disciplinary expertise and AI familiarity consistently perceived snippets—regardless of AI involvement—as clearer and more compelling. Responsible and transparent use of GenAI can improve research manuscripts without compromising reviewers' perceptions. *Second*, we report how our survey revealed reviewers' contradictory perceptions of AI and human authorship indicators. This revelation has substantial implications for fair and unbiased manuscript evaluation with the potential to reshape peer-review processes across disciplines. We encourage authors to prioritize manuscript coherence, research validity, and effective communication, without letting their attitudes and misconceptions about GenAI influence their assessments. *Third*, we show that reviewers valued the subjective expressions of human authors in research manuscripts. This "human touch" resonated with reviewers because it maintains the collaborative nature of the research community. Therefore, we suggest researchers retain adequate involvement in their writing and act as the primary driver of the writing process—even with GenAI assistance. *Fourth*, our qualitative findings show that reviewers' apprehensions about GenAI may worsen the publish-or-perish culture in academia. This could disproportionately affect researchers who rely on traditional writing methods. As a result, it would ultimately stifle human creativity. Our findings directly inform best practices for integrating GenAI in manuscript preparation—while maintaining research integrity—because we identify specific elements that shape reviewers' perceptions. We conducted this research to provide crucial insights for the timely development of ethical AI use policies in academia. In addition, our findings contribute to the ongoing debate about GenAI's role in academia by providing empirical evidence of its effects on peer review—a hallmark of scientific progress.

## 2. Background and related work

In this section, we summarize the technical evolution of GenAI as a manuscript writing assistant and the emerging perceptions and concerns within the academic community. In the end, we illustrate how our research addresses these concerns and promotes the ethical, transparent, and effective use of GenAI to support future researchers.

### 2.1. *Generative AI as a writing assistant*

Manuscript writing is crucial for researchers to share their ideas and contribute to their fields. However, writing high-quality research papers is challenging due to the need to simplify complex findings while ensuring accuracy, logical flow, and adequate evidence (Gupta, Jaiswal, Paramasivam, & Kotecha, 2022). Beginners and non-native English speakers often struggle with using proper terminology and literature references (Gupta et al., 2022; Inouye & McAlpine, 2019; Liang et al., 2023). In addition, manuscript writing often competes with other responsibilities like teaching and supervising (De Rond & Miller, 2005), making efficiency and time management vital. The pressure of "publish or perish" mindset (De Rond & Miller, 2005) further intensifies these challenges. GenAI thus become valuable in research writing to ease researchers' burden on writing and help them keep their focus on the innovative and critical aspects of their research.

With the rise of Large Language Models (LLM), GenAI's potential to transform manuscript writing has garnered significant interest (Mohamed et al., 2024; Som, 2023; Tu et al., 2024). Traditional writing assistants offer word and sentence corrections, synonym suggestions, and sentence completion predictions (Arnold, Gajos, & Kalai, 2016; Chen et al., 2019; Quinn & Zhai, 2016). In contrast, GenAI offers a broader array of functionalities to ensure high-quality writing across diverse research disciplines, such as inspiring new ideas (Lee, Liang, & Yang, 2022; Shaer, Cooper, Mokryn, Kun, & Shoshan, 2024), enhancing readability (Babl & Babl, 2023), and assisting with narrative construction and creative writing (Lee et al., 2022; Singh, Bernal, Savchenko, & Glassman, 2023; Yuan, Coenen, Reif, & Ippolito, 2022). However, GenAI has the limitation of generating factually incorrect information, known as hallucination (Achiam et al., 2023; Ji et al., 2023). For example, researchers have reported encountering fake references from GenAI (Committee on Publication Ethics [COPE], 2023a). In addition, GenAI can be opinionated, which influence researchers' perspectives and attitudes conveyed in the writing and compromise research integrity (Jakesch, Bhat, Buschek, Zalmanson, & Naaman, 2023). Therefore, while GenAI holds benefits for manuscript writing, its use requires researchers' careful consideration to avoid the risks.

These problems highlight the importance of transparently disclosing the use of GenAI. Such disclosure enables reviewers and readers to critically evaluate the research, be aware of potential biases or inaccuracies introduced by GenAI. Our study investigates reviewers' perceptions and misconceptions, reduces current concerns and hesitations among researchers, encourages researchers to openly disclose their GenAI use, and fosters a more transparent and accountable research environment.

### 2.2. Perceptions of generative AI in research community

A central debate in the research community regarding GenAI involves authorship and content attribution (COPE, 2023b). Research manuscripts reflect the knowledge, expertise, and contributions of its author researchers (The Committee on Publication Ethics [COPE], 2019). The use of GenAI in manuscript writing has raised questions about how to acknowledge its involvement, as crediting it as a co-author is inappropriate because "AI tools cannot meet the requirements for authorship as they cannot take responsibility for the submitted work" (COPE, 2023b), para. 2). GenAI also cannot be accountable for the content it produces (COPE, 2023a; COPE, 2023b). Beyond authorship, ethical concerns arise, such as copyright infringement from using third-party materials, possible conflicts of interest, and plagiarism issues that replicate contents and images, ideas, and methods from already published works (COPE, 2023a; Lund et al., 2023). In 2023, the Committee on Publication Ethics (COPE) recommended that authors explicitly disclose the use of AI-assisted technologies, including LLMs like ChatGPT, in their work (COPE, 2023b). Following COPE's lead, the Association for Computing Machinery (ACM) established policies on GenAI, stating "the use of generative AI tools and technologies to create content is permitted but must be fully disclosed" (Association for Computing Machinery [ACM], 2023). Following these, efforts are made to develop comprehensive reporting guidelines for evaluating the impact of tools like ChatGPT on scientific research writing, as seen in initiatives by Elsevier. and the World Association of Medical Editors (2023). These guidelines aim to promote transparency by providing a framework for declaring the use of GenAI in research.

Scholarly work revealed two opposing perceptions of AI-generated content: algorithm aversion and algorithmic appreciation. *Algorithm aversion* is a negative bias towards AI-generated content, even when the AI output is objectively better than human-produced content (Burton et al., 2020; Hong, 2018). For example, people tend to rate AI-written content as inaccurate regardless of its truthfulness (Longoni et al., 2022). In addition, informing users about AI involvement can harm the creator-reader relationship rather than facilitate content judgment (Irene, 2024). This bias worsens after seeing AI makes mistakes (Berkeley et al., 2015). On the other hand, *algorithmic appreciation* refers to when people are more willing to adhere advice from an algorithm over a human (Logg et al., 2019), and find AI-created articles more credible with higher journalistic expertise (Graefe et al., 2018).

Manuscript writing involves various decisions about word choice and sentence structure to effectively convey authors' meaning and purpose, with each word representing a decision made by the authors (Max, 2024). With GenAI, many of these decisions are delegated to AI, which relies on highly probable options, pre-defined rules, large databases, or specific text corpora (Max, 2024). This delegation can reduce human authors' sense of ownership (Fiona et al., 2024; Lee et al., 2022), which may potentially lead to irresponsible assertions in research papers. Therefore, regulating the extent of GenAI assistance is crucial for maintaining the accountability and credibility of research publications. Our research aims to encourage transparency in disclosing GenAI use, which is the foundational step for responsible AI augmentation in research manuscript writing.

### 2.3. Connection to our research

While guidelines exist to guide researchers and promote transparency in research community, many researchers are hesitant to acknowledge their use of GenAI in their manuscripts (e.g., LinkedIn.com, 2024; Reddit.com, 2024; Retraction Watch; Twitter.com, 2023). Although previous studies have examined human ability to detect AI-generated content (e.g., Gao et al., 2023; Köbis & Mossink, 2021; Ragot, Martin, & Cojean, 2020), these studies were not conducted in the context of research publications and were not conducted with participants with experience reviewing academic manuscripts in peer-reviewed venues. Therefore, their findings offer limited insight into the specific issue of GenAI use in research manuscript writing. Our study addresses this gap by investigating experienced reviewers' perceptions and misconceptions on manuscripts due to GenAI use. Through this investigation, we aim to reduce researchers' concerns about negatively impacting reviewers' perceptions and judgments, and encourage them to openly acknowledge their use of GenAI in future manuscripts. Given the increasing adoption of GenAI in research writing and the ethical needs of research transparency, our research is crucial and urgent in charting a path for an ethical and beneficial GenAI augmentation in research manuscripts writing while avoiding detrimental consequences.

## 3. Methodology

To investigate reviewers' perceptions of GenAI use in research writing, we employed a text snippet-based online survey. After obtaining Research Ethics Board approval [REB protocol #46328], we recruited 17 participants who have experience reviewing manu-scripts for publication at top-tier HCI conferences, including CHI[3] and CSCW.[4] We refer to our participants as "reviewers" in the following sections. Reviewers were presented with six snippets tailored to their areas of expertise in HCI, chosen from 16 example human-written ab-stracts and 32 GenAI-augmented snippets. The six snippets were pre-sented in a randomized sequence. This approach allowed us to explore reviewers' perception on a wide range of topics with different levels of GenAI use without overwhelming them with a long survey. In this section, we describe our snippet design, survey development, participant recruitment, and data analysis procedure.

### 3.1. Study material construction

In research paper writing, GenAI is used in various ways from recommending texts, perform spelling or grammar corrections, to generating entire sections (ACM, 2023). To comprehensively evaluate reviewers' perception, we present each participant with three types of snippets (Content_Type).

1. *Original*: snippets written entirely by human authors.
2. *Paraphrased*: snippets rephrased with a GenAI by rewriting human-written text while preserving its original meaning.
3. *Generated*: snippets generated entirely with a GenAI by using human-written text as reference to ensure relevance to the original manuscript.

In this section, we discuss the selection of original human-written snippets, and the production of paraphrased and generated snippets

[3] The ACM CHI Conference on Human Factors in Computing Systems (CHI).

[4] The ACM Conference on Computer-Supported Cooperative Work and Social Computing (CSCW).

using GenAI prompts.

#### 3.1.1. Original snippets

To ensure the comprehensive coverage of our *original* snippets, we selected abstracts from example papers from submission topics of CHI 2023 conference,[5] the premier venue for HCI research.[6] For each topic, we selected the most-cited paper published before the prevalent use of GenAI in November 2022 to ensure it was written by human researchers. When multiple papers had the same citation numbers, we subsequentially selected papers based on download counts and the most recent publication date. This process resulted in a total of 16 abstracts as our *original* snippets. Details of these source papers are in Appendix section C.

We chose to use abstracts due to three considerations. First, abstracts are crucial for research manuscripts as they comprehensively summarize the papers' significance, research goals, methodology, findings, and contributions (Laura Belcher, 2019). Second, in early stage of a peer-review process, abstracts guide editors and reviewers in efficiently evaluating a manuscript (Laura Belcher, 2019). Third, since we recruit experienced reviewers who are academia and industry professionals, using abstracts ensures our study is manageable and not overly time-consuming while still offering sufficient information for evaluating participants' perceptions.

#### 3.1.2. Paraphrased and generated snippets

The selected *original* snippets were then processed through GenAI—Google Gemini[7]—to create the corresponding *paraphrased* and *generated* snippets. We chose Gemini for its ability to provide comprehensive summaries, valuable suggestions, and rationales, as well as its transparency in disclosing limitations rather than fabricating content, which distinguish it from other GenAI tools such as ChatGPT (Tu et al., 2024).

Building upon literature on constructing GenAI prompts (Mollick & Mollick, 2023) and discussions with our research team of GenAI researchers and enthusiasts, we incorporated four components in our construction of the prompts for snippets processing.

(1) *Goal:* the goal of the prompt. For producing *paraphrased* snippets, we set the goal as "rephrase" the original snippet; For producing *generated* snippets, we set the goal as "improve" the paraphrased snippet to allow GenAI to maximize its creativity while ensuring the content consistency.
(2) *Step-by-step instruction:* the detailed instruction that specifies expected GenAI behaviour step-by-step. For producing *paraphrased* snippets, we provided a guide based on best practices of abstract writing (Laura Belcher, 2019). For producing *generated* snippets, we used two sequential prompts that guide GenAI to first generate a new snippet based on the paraphrased snippet and the introduction section of the paper, then refine its contribution statements based on the manuscript's conclusion section.
(3) *Context:* the context information that facilitates the GenAI behaviours. For producing *paraphrased* snippets, the original snippet served as the context. For producing *generated* snippets, the paraphrased snippet and the corresponding manuscript's introduction and conclusion sections were used.
(4) *Constraints:* to ensure consistency in length, we set a 150-word constraint for both *paraphrased* and *generated* snippets based on typical CHI submissions.[8]

Researchers in our team reviewed the snippets to ensure consistency in content and length across the three content_types. Figs. 1 and 2 illustrate the prompt structure, and Appendix section D provides examples of the snippet production process in Gemini. This approach ensures that the snippets derived from the same abstract maintain consistent length, level of detail, and content. In this way, we ensures that our reviewers assess the snippets based on variations in writing style, word choice, structure, and flow due to GenAI involvement, rather than differences in interpretations and opinions that naturally vary among human authors.

### 3.2. Survey design

In this section, we provide a detailed description of our survey design. Fig. 3 summarizes the survey flow. A complete set of questions is included in Appendix section E.

#### 3.2.1. Screening questionnaire

The survey began with a study information sheet and consent form, followed by a screening questionnaire. Our screening targeted participants who have experience serving as reviewers in peer-reviewed HCI conferences. Participants had to be at least 18 years old, have previous experience as a reviewer or associate chair, and have encountered or suspected the undisclosed use of GenAI in submissions they reviewed.

#### 3.2.2. Instruction and presentation of snippets

To ensure reviewers' perceptions were related to their experience with GenAI, not conventional writing assistants, we first provided a description of GenAI's functionality: "AI writing assistants can help researchers by suggesting phrasing, structuring sentences, and even generating initial drafts." Reviewers then selected two research topics from the 16 CHI'23 topics (see Q1 & Q2 in Appendix section E)—one in which they were most knowledgeable and one in which they had the least knowledge. From each topic, we presented the original, AI-paraphrased, and AI-generated snippets from an example paper (as described in subsection 3.1.1). This approach allowed us to compare reviewers' perceptions and judgements varied between content_types, and investigate how their expertise influenced their perceptions. To avoid biasing reviewers, we did not disclose the content_type of each snippet. We described the six snippets as could be human-written or AI-processed without confirming AI or human authorship. The three snippets from the same abstract were presented in random order. Since the snippets were from published papers, we included a bold red text instructing reviewers not to search for the snippets in literature databases.

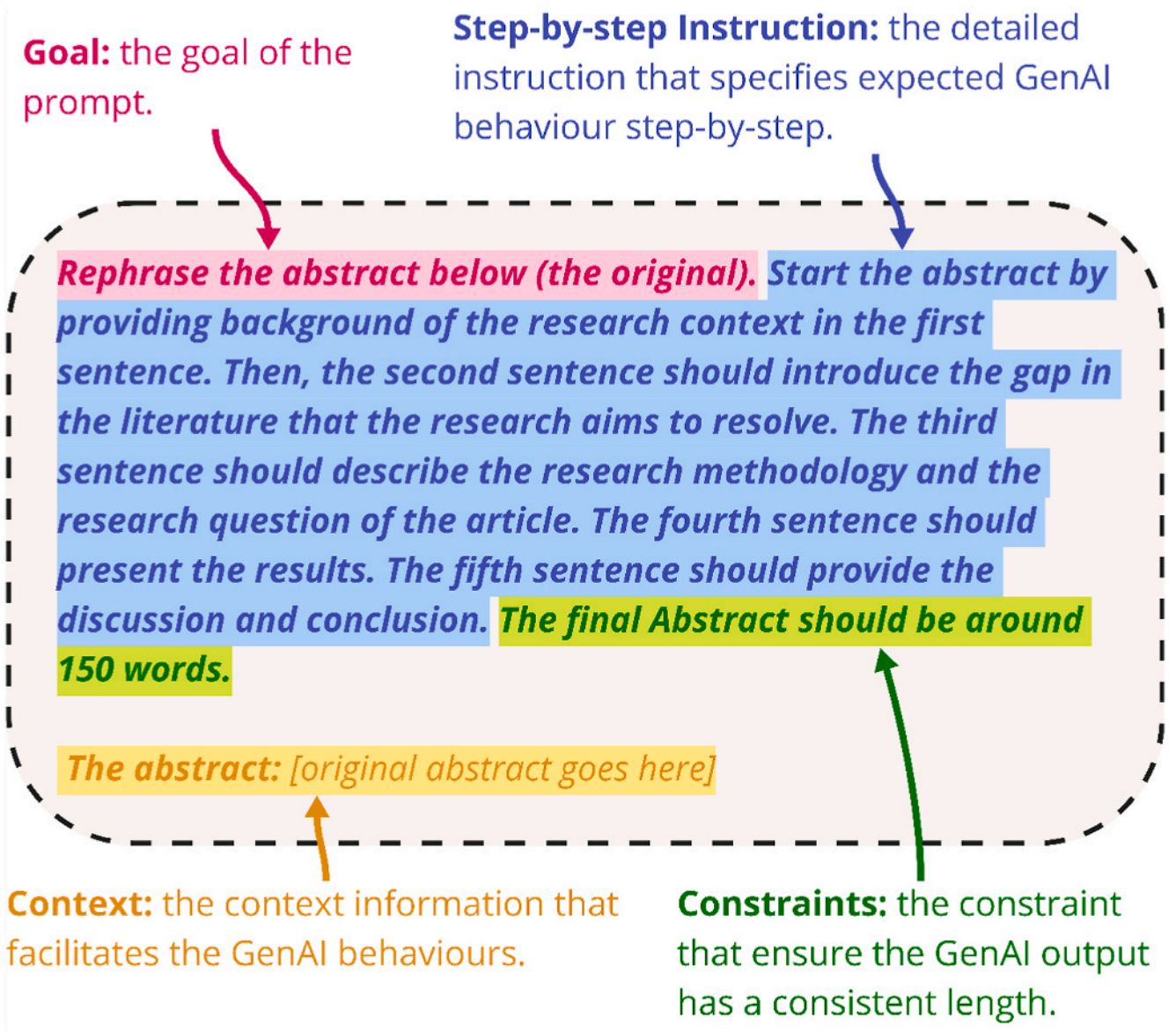

**Fig. 1.** Example prompt used for creating a *paraphrased* snippet from the original snippet.

[5] CHI'23. "Selecting a Subcommittee". Last modified (n.d.). Last accessed on March 19, 2024. https://chi2023.acm.org/subcommittees/selecting-a-subcommittee/.

[6] As of June 7, 2024, CHI was ranked as the premier venue in human-computer interaction research, with h5-index at 122, twice of the venue ranked as the second. See: https://scholar.google.ca/citations?view_op=top_venues&hl=en&vq=eng_humancomputerinteraction.

[7] Google Gemini. https://gemini.google.com/app.

[8] CHI 2023 | Papers. See section "Preparing and Submitting Your Paper" on https://chi2023.acm.org/for-authors/papers/.

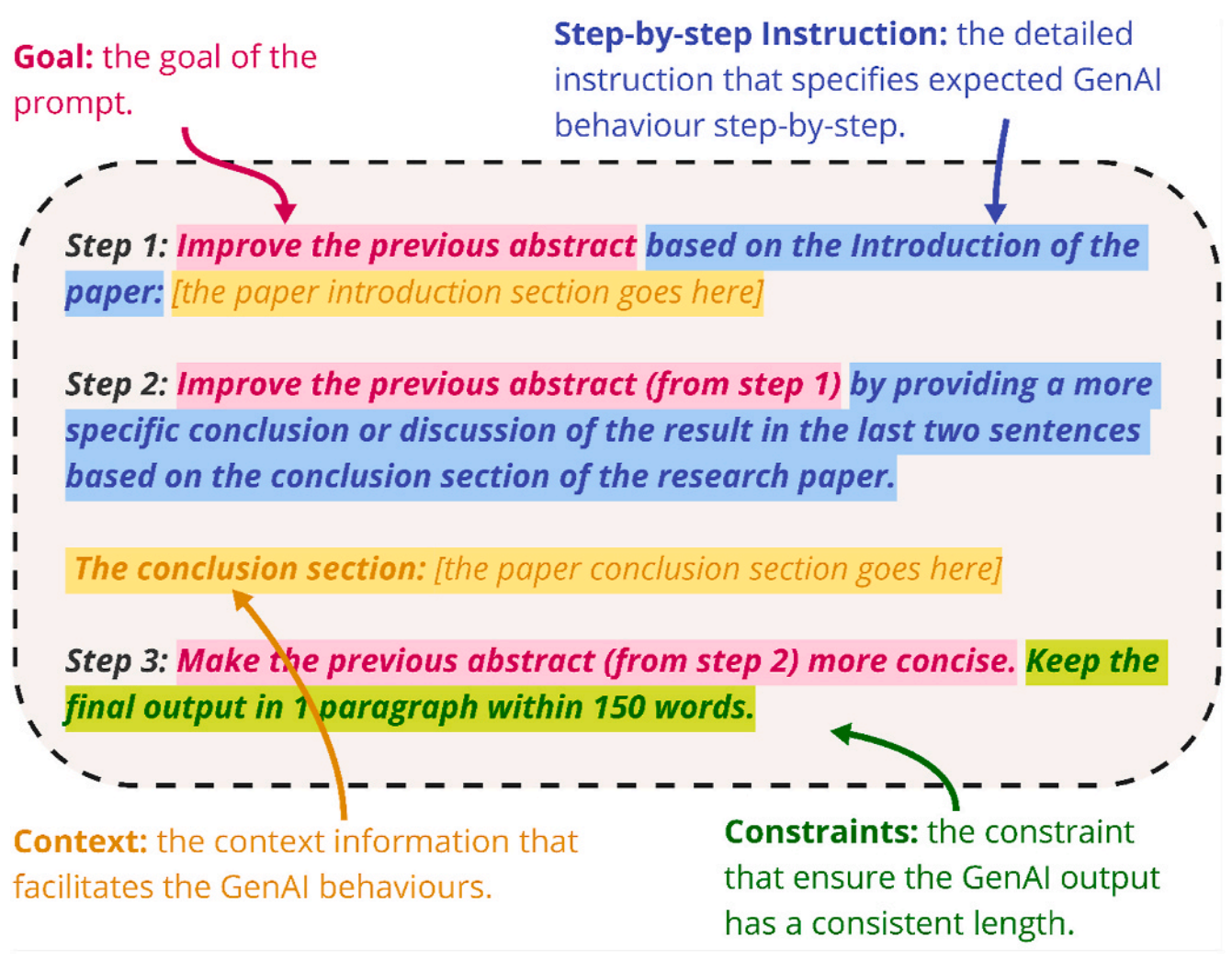

**Fig. 2.** Example prompt used for creating a *generated* snippet from the paraphrased snippet, and the manuscript's introduction and conclusion sections.

### *3.2.3. Perceptions of the snippets and the research presented*

For each snippet, reviewers were asked to provide a more detailed rating of their expertise in the topic, using a scale from *0—no knowledge or expertise in this field* to *10—I am an expert in this field*. We coded these responses as disciplinary_expertise in our statistical analysis. This question served three purposes. First, it clarified what "the most" and "the least" knowledgeable meant by each reviewer. Second, it captured cases when reviewers misidentify that a paraphrased or generated snippet is from a completely different abstract than its original. Third, it acted as an attention check. Reviewers selecting a topic they claimed to be most or least knowledgeable in but giving an opposite rating here indicated a lack of attention to our instructions.

To determine if reviewers' judgements on research integrity, value, and soundness varied because of the writing across the three content_types, we asked them to rate each snippet's accuracy (perceived_accuracy), reliability (perceived_reliability), honesty (perceived_honesty), clarity (perceived_clarity), and compellingness (perceived_compellingness) in representing the research (International Center for Academic Integrity [ICAI], 2018). Reviewers rated these aspects on 5-point Likert scales, from *1—strongly disagree* to *5—strongly agree*, following Longoni et al. (Longoni et al., 2022)'s study on readers' perception of news-headlines.

Next, we asked reviewersto rate their perceived level of AI involvement (perceived_AI_involvement) in each snippet's writing process on a scale from *0—completely human* to *10—completely GenAI*, inspired by the methodology from Draxler et al. (Fiona et al., 2024), which asked participants to select the possible author attribution from a set of randomized options. Our 10-point scale offered finer granularity for reviewers to express their perceptions more accurately. For reviewers who suspected at least some degree of GenAI involvement (i.e., not completely human written), we included a highlight question, asking them to highlight specific sentences they believed were AI-processed. After that, reviewers were asked to share observations about the snippet's style, structure, or content that influenced their perception of its authorship on an open-ended question. The combination of these questions allowed us to identify specific segments that influenced reviewers' judgments.

To ensure data quality, we included an attention check question between the six snippets. The question asked reviewers to select a specific option. Reviewers who failed to select the designated option were excluded from our analysis for not following instructions.

### *3.2.4. General perception of GenAI and demographic information*

After all six snippets, we closed the survey with questions about reviewers' general perceptions of GenAI writing. We asked about their views on the capability of human researchers (perceived_human_researcher_capability) and GenAI in communicating research ideas and outcomes through writing (perceived_AI_capability). These questions aimed to assess the reviewers' algorithmic aversion or appreciation (Burton et al., 2020; Graefe et al., 2018; Hong, 2018), as their negative or positive attitudes toward GenAI may influence their perceptions of the snippets. Finally, we asked reviewers about their demographic information, estimated the number of papers they had reviewed (peer-review_experience), and use of GenAI in their own writing (AI_familiarity). We included these questions because AI background knowledge can influence perceptions (Ehsan et al., 2024), and people's algorithmic aversion increases after witnessing AI mistakes (Berkeley et al., 2015). Reviewers were also given an open-ended space for additional comments on our study before completing the survey.

## *3.3. Participants recruitment and demographics*

Before distributing the survey, we piloted the questionnaire with five

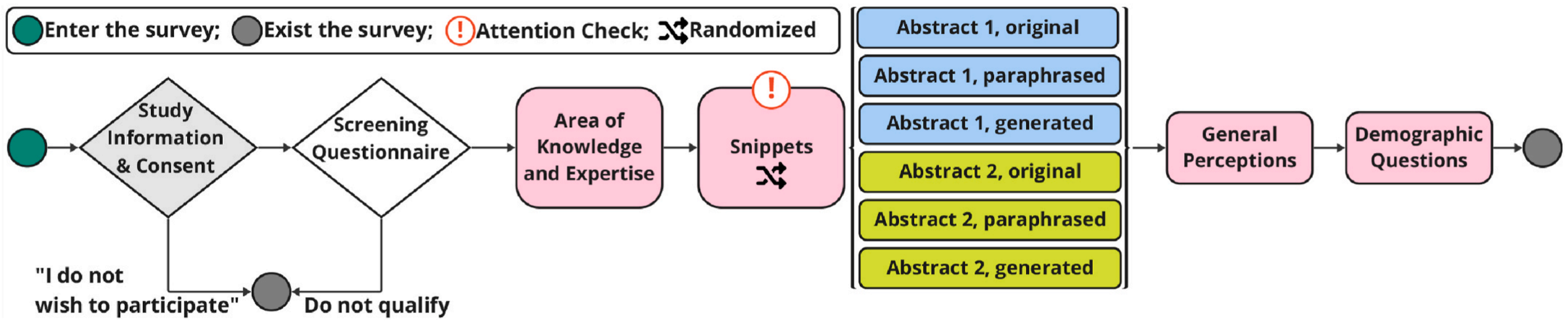

**Fig. 3.** Survey flow: After passing the screening questionnaire, N = 17 reviewers selected their areas of expertise. They were then shown six snippets, including three content_types of two abstracts selected from a total of 16 original snippets based on their reported expertise. For each snippet, we assessed reviewers' perception and judgment of the content and the research presented. Finally, reviewers shared their general views on GenAI in research writing and provided demographic information.

PhD students with peer-review experience and refined the language and question structure based on their feedback to improve clarity, comprehension, and conciseness. A prior power analysis (Faul et al., 2007, 2009) for a within-subject Wilcoxon signed-rank test determined that a sample size of $N = 15$ was needed, with an effect size = 0.8, a power = 0.8, and a margin for random error$\leq$ 5%. Following ethics approval, we recruited participants using a snowball sampling method in April and May 2024. Our research team reached out to CHI and CSCW conference committees for participation and assistance in distributing recruitment materials. This recruitment method was used due to the difficulty in recruiting reviewers, even in real peer-review process (Henderson et al., 2020, pp. 957–959). We closed the survey on May 7, 2024, one month after receiving the last response, resulting in a total of 41 responses. Of these, we excluded 23 responses for completing less than 50% of the questions (11 only completed the consent form) and one for failing the attention check. Our final analysis was based on the remaining $N = 17$ valid responses.

Our study included 17 reviewers from premier HCI conferences, who represent a range of experience levels and areas of expertise within the field. While our sample size is limited, it embraces diverse perspectives, including novice and senior reviewers. The varied backgrounds of our participants in HCI sub-fields—with Games and Play being the most common expertise area—provide valuable insights into reviewer perceptions. However, we acknowledge that this sample may not be completely representative of the entire HCI reviewer community. Despite this apparent limitation, our findings offer crucial insights into reviewer attitudes towards AI-augmented writing in HCI. Table 1 summarizes the demographics of our 17 reviewers, including 53% women, 41% men, and one (6%) non-binary. Most reviewers were aged 27 to 49 and held post-secondary degrees (graduate or professional = 94%, bachelor = 6%), with a job occupation of academic researcher (graduate researchers = 24%, postdoctoral researchers = 29%, and professors = 35%). The reviewers included novice and senior reviewers with varied areas of expertise, with Games and Play being the most selected topic (35%). In terms of personal GenAI use, 59% of reviewers reported sometimes using it for targeted research writing purposes, 12% rarely used it, and 29% had never used it.

### 3.4. Data analysis

We present our quantitative data analysis and corresponding results in section 4. For the qualitative open-ended question, we conducted an inductive thematic analysis with two researchers, following the established guideline by Clarke et al. (Clarke, Braun, & Hayfield, 2015). The two researchers began by reviewing the data to familiarize themselves with it, and ensure there were no blank or incoherent responses to each question. We retained "N/A" responses, which represent an inability to differentiate human-written snippet from GenAI output. During this process, both researchers took rough notes on the trends they spotted in the data. Next, the two researchers independently coded the same 15% ($n = 16$) of the total responses ($N = 102$) then held a meeting to discuss and resolve disagreements in $n = 3$ responses. In the same meeting, the researchers collaboratively reviewed the same 16 responses to not neglect any insights, and refined and added to the codes. After the meeting, an initial set of themes was developed based on the trends emerged in the codes, which formed the first version of our codebook. In the second iteration, the two researchers again independently coded the same 42% ($n = 43$) of responses of the remaining responses. They then held another meeting to resolve disagreements ($n = 7$ responses) and collaboratively reviewed the same responses to check all insights were captured, and further refined the codes, themes, and codebook. In the third iteration, the researchers repeated this process with the remaining 42% ($n = 43$) of responses and discussed and resolved disagreements ($n = 3$ responses) in a meeting. As part of this meeting, they collaboratively reviewed the same responses, and refined and added to the existing codes, themes, and codebook. Finally, an additional discussion session was held to review all codes holistically and identify broader patterns across the dataset. After this session, we finalized our themes and codebook. Throughout our thematic analysis process, we did not calculate inter-coder reliability, because it "prioritises uniformity over depth of insights" and often results in superficial themes, especially for studies with more than 20 codes (like ours) (Clarke et al., 2015, p. 303). Instead, the two researchers discussed and resolved disagreements in the regular meetings. We present our final codebook and themes in Appendix section A.

## 4. Findings

### 4.1. RQ1: How much are reviewers aware of the use of AI in the context of research writing?

Table 2 shows the response distribution among the $N = 17$ reviewers regarding their perceived_AI_involvement across the three content_types. Both original human written snippets and AI-generated snippets received a median = 5, with a mean = 4.44 (SD = 3.13) and mean = 5.12 (SD = 3.18), respectively. This result indicates that reviewers generally believed GenAI was similarly involved in both human-written and AI-generated snippets. This similarity revealed a general misconception about GenAI use in snippets and suggested the difficulty in differentiating between AI-generated and human-written snippets among reviewers. Compared to these, the rating for AI-paraphrased snippets is notably lower (median = 2, mean = 2.74, SD = 2.61).

To validate the observed differences in reviewers' perceptions, we performed a Friedman test (Friedman, 1937) and confirmed significant within-subject differences across the three types of snippets ($\chi^2 = 6.92$, $df = 2$, $P = 0.03$). We further conducted post-hoc pairwise Wilcoxon comparisons (Robert, 2007, p. 3) with Bonferroni correction (Chen, Feng, & Yi, 2017) (see Table 2). The result shows that, compared to AI-generated snippets, reviewers perceived significantly lower AI involvement in AI-paraphrased snippets ($W = 92$, $P = 0.01$, $r = -0.60$). There was no significant difference in reviewers' perceptions between AI-generated and human-written snippets ($W = 80$, $P = 0.55$). Additionally, no significant difference was found between reviewers' perceptions of human-written and AI-paraphrased snippets ($W = 26.5$, $P = 0.06$). The validity of these results are further supported by our reviewers' qualitative responses, with several of them indicated they were confused about which snippets were AI- or human-written.

### 4.2. RQ2: How much is reviewers' judgement of research and manuscript influenced by the use of AI in its writing?

Table 3 presents the distribution of reviewers' judgments across the three content_types. The result shows that reviewers' responses were mainly neutral ($mean = 3.29$, $SD = 1.12 \sim mean = 3.82$, $SD = 0.80$), and there is no sizeable differences between reviewers' perception on the accuracy, reliability, honesty, clarity, and compellingness.

To further validate our observations, we conducted a Friedman test (Friedman, 1937) and found no significant within-subject differences in reviewers' perception across the three content_types. We suspect that this result is because our reviewers neither exhibited algorithmic aversion nor appreciation, but had neutral opinion towards GenAI. To validate this, we conducted a within-subject Wilcoxon signed-rank analysis (Robert, 2007, p. 3) with Bonferroni correction (Chen et al., 2017) to compare reviewers' perceived_human_researcher_capability (mean = 4.35, SD = 0.79) and perceived_AI_capability (mean = 4.06, SD = 0.97). The results showed no significant difference in reviewers' perceptions of AI and human researchers' writing abilities (W = 31.5, P = 0.28, r = −0.26). Although the effect size is small, the validity of this result is supported by our reviewers' lower perceived AI involvement and higher perceived honesty in AI-paraphrased snippets in subsection 4.3 and their qualitative responses that highlighted the advantages and weaknesses

**Table 1**
Participant reviewers' ($N = 17$) Demographic Information.

| Age | | Occupation | | Area of Expertise[a] | | AI Familiarity | |
|---|---|---|---|---|---|---|---|
| Range | 27–49 | Professor | 6 (35%) | Games and Play | 6 (35%) | Sometimes | 10 (59%) |
| Mean | 34.52 | Postdoctoral Researcher | 5 (29%) | Interaction Techniques & Modalities | 3 (18%) | Rarely | 2 (12%) |
| SD | 5.62 | Graduate Researcher | 4 (24%) | Design | 2 (12%) | Never | 5 (29%) |
| | | Industry Professional | 1 (6%) | Learning, Education, and Families | 2 (12%) | | |
| **Gender** | | Other-freelancer | 1 (6%) | Critical Computing, Sustainability, and Social Justice | 1 (6%) | **Peer-review Experience** | |
| Woman | 9 (53%) | **Education Level** | | Health | 1 (6%) | Range | 5–500 |
| Man | 7 (41%) | Graduate or professional | 16 (94%) | Specific Applications Areas | 1 (6%) | Mean | 110.94 |
| Non-binary | 1 (6%) | Bachelor | 1 (6%) | Understanding People | 1 (6%) | SD | 152.93 |

*Note.*
[a] Research areas are based on the topics from the ACM CHI Conference on Human Factors in Computing Systems in 2023 (CHI'23) subcommittees.

**Table 2**
Reviewers' (N = 17) perceived AI Involvement (0-completely human to 10-completely AI) Across Content Types.

| Perceived AI Involvement | n | Median | Mean | SD | Min | Max |
|---|---|---|---|---|---|---|
| ***Content Type*** | | | | | | |
| Original | 17 | 5 | 4.44 | 3.13 | 0 | 9 |
| Paraphrased | 17 | 2 | 2.74 | 2.61 | 0 | 10 |
| Generated | 17 | 5 | 5.12 | 3.18 | 0 | 10 |

| Friedman Test | | Post-hoc Pairwise Wilcoxon tests[a] | | |
|---|---|---|---|---|
| Friedman chi-squared | 6.92 | ***Content Type*** | Original | Paraphrased |
| Df | 2 | Original | | |
| P | 0.03 | Paraphrased | W = 13, P = 0.06, r = −0.46 | |
| | | Generated | W = 93.5, P = 0.55, r = −0.14 | W = 92, P = 0.01, r = −0.60 |

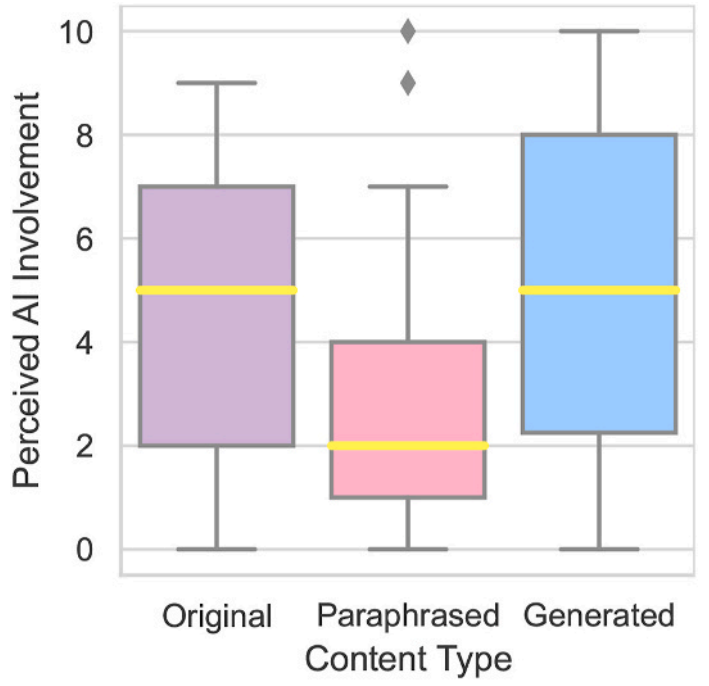

Note.
[a] *P adjusted with Bonferroni correction. SD=Standard Deviation, W=test statistic, r=effect size.*

from both AI and human writing in subsection 4.4.

### 4.3. RQ3: To what extent do reviewers' peer-review experience, disciplinary expertise, and AI familiarity influence their perception and judgement?

In this section, we evaluate how factors including content_type, reviewers' disciplinary_expertise, AI_familiarity and peer-reviewer_experience influence their perceived_AI_involvement and judgements on the manuscript and presented research. We used Cumulative Link Mixed Model (CLMM) regression[9] and included participant identifiers as random effects. CLMM is well-suited for repeated measures experiments with ordinal dependent variables, as in our study where reviewerswere presented with multiple snippets in parallel (Christensen, 2019). We conducted a series of Multivariate CLMM regressions, using reviewers' perceived_AI_involvement, perceived_accuracy, perceived_reliability, perceived_honesty, perceived_clarity, and perceived_compellingness as the dependent variable (DV) and the factors as the predictors. Table 4 shows the final models with predictors ranked by their contribution to the DV, determined by the global minimum Akaike Information Criterion (AIC) (Kadane & Lazar, 2004) values obtained upon adding each predictor. Predictors with the highest contribution (lowest AIC) are ranked first.

As shown in Table 4, the results revealed relationships between reviewers' perceived_AI_involvement and the predictors content_type and AI_familiarity, with content_type had the greatest contribution. Specifically, reviewers perceived significantly lower AI involvement in AI-paraphrased snippets compared to original human-written snippets. This result extends our within-subject comparison in subsection 4.1. In addition, reviewerswho rarely used AI in their research writing (AI_familiarity) perceived lower AI involvement than those who never used it, indicating that even minimal AI familiarity influences perceptions of AI involvement.

Furthermore, reviewers' perceived_honesty, perceived_clarity, and perceived_compellingness showed significant positive associations with their disciplinary_expertise, AI_familiarity and content_type, with the disciplinary_expertise had the greatest contribution. Specifically, reviewers with greater expertise in the relevant research field perceived higher levels of honesty and clarity, particularly in AI-paraphrased snippets compared to original human-written ones. Moreover, we found that reviewers' *AI_familiarity* positive associated with their perceived compellingness. That is, reviewers who sometimes used GenAI in their writing found the snippets more compelling than those who never used GenAI. Reviewers who sometimes used GenAI in their writing perceived higher level of clarity than those who never did. These results' validity are further supported by our qualitative findings where revie-wers appreciated well-structured sentences and good readability in snippets from GenAI (see subsection 4.4).

### 4.4. RQ4: What aspects of research writing impact reviewers' perception and judgement?

In this section, we discuss the themes derived from reviewers' qualitative responses (see Fig. 4). For clarity, thematic analysis themes are in *italics*, and reviewers' quotes are in *"italicized quotations."* The survey question is detailed in Appendix section E (Q9). We discuss how these themes are related to our quantitative findings in section 5.

Our thematic analysis of $N = 102$ open-ended responses revealed five major themes that influence reviewers' perception of the author of snippets: 1) Writing and Sentence Structure, 2) Word Choice, 3) Problematic Statement, 4) Expression, and 5) Carefully Crafted Statement. Interestingly, the codes under these themes revealed reviewers' contradictory opinions, which aligned with our quantitative findings that reviewers struggled to differentiate AI-generated snippets from those written by human researchers (see subsection 4.1). Our focus was on the reviewers' reasoning behind their perceptions. Therefore, our thematic

[9] We used the Ordinal R-package (https://cran.r-project.org/web/packages/ordinal/).

**Table 3**
Reviewers' (N = 17) Perceived Content Quality (0-completely disagree to 5-completely agree) Across Content Types.

| **Perceived accuracy** | **n** | **Median** | **Mean** | **SD** | **Min** | **Max** | **Friedman Test**[a] | |
|---|---|---|---|---|---|---|---|---|
| Original | 17 | 3 | 3.35 | 0.95 | 1 | 5 | Chi-squared | 2.48 |
| Paraphrased | 17 | 4 | 3.68 | 0.84 | 2 | 5 | df | 2 |
| Generated | 17 | 3 | 3.29 | 1.12 | 1 | 5 | P | 0.29 |
| **Perceived reliability** | | | | | | | | |
| Original | 17 | 3.5 | 3.50 | 1.02 | 1 | 5 | Chi-squared | 1.85 |
| Paraphrased | 17 | 3.5 | 3.65 | 0.81 | 2 | 5 | df | 2 |
| Generated | 17 | 3 | 3.32 | 1.09 | 1 | 5 | P | 0.40 |
| **Perceived honesty** | | | | | | | | |
| Original | 17 | 3 | 3.35 | 0.85 | 1 | 5 | Chi-squared | 4.68 |
| Paraphrased | 17 | 4 | 3.82 | 0.80 | 3 | 5 | df | 2 |
| Generated | 17 | 3 | 3.38 | 1.04 | 1 | 5 | P | 0.10 |
| **Perceived clarity** | | | | | | | | |
| Original | 17 | 4 | 3.47 | 1.11 | 1 | 5 | Chi-squared | 0.35 |
| Paraphrased | 17 | 4 | 3.74 | 0.96 | 2 | 5 | df | 2 |
| Generated | 17 | 4 | 3.56 | 1.21 | 1 | 5 | P | 0.84 |
| **Perceived compellingness** | | | | | | | | |
| Original | 17 | 3 | 3.30 | 0.98 | 1 | 5 | Chi-squared | 0.82 |
| Paraphrased | 17 | 4 | 3.52 | 0.94 | 2 | 5 | df | 2 |
| Generated | 17 | 4 | 3.45 | 1.06 | 2 | 5 | P | 0.66 |

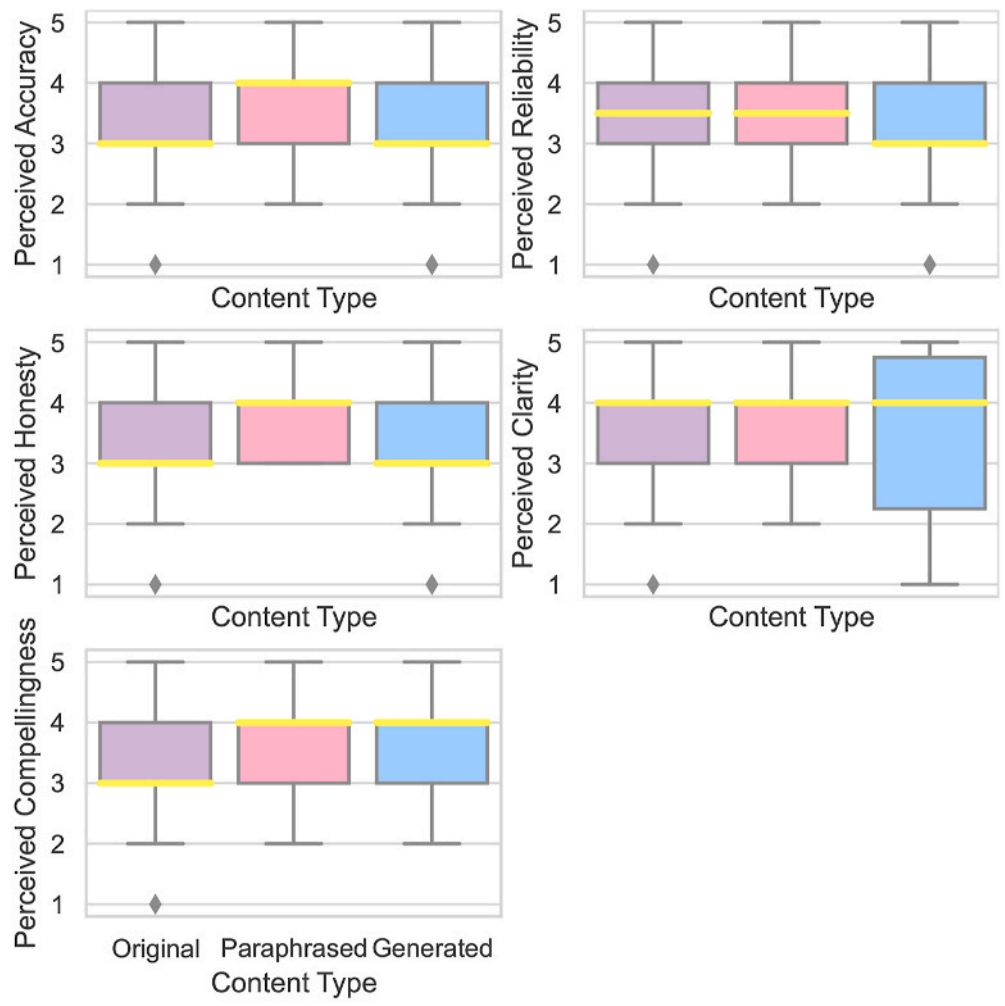

Note.
[a] *P adjusted with Bonferroni correction. df=degrees of freedom. Cronbach's alpha=0.89.*

analysis approach and our presentation of the codes and themes in this section are not guided by the content types but rather by the aspects within the snippets that shaped reviewers' perceptions.

#### 4.4.1. *Theme 1: Writing and Sentence Structure*

A primary perception among the reviewers was that they often associate the snippets with problems of *incoherent logic and phrasing*, with illogical transitions, unclear flow, and misuse of field-specific terminologies being AI produced (in 27% of responses). In contrast, 12% of responses noted that a *human produces neatly edited sentences*, and 2% mentioned that experienced researchers know how to structure sentences effectively (*trained researchers know the writing structure*). Moreover, 1% highlighted that humans tend to write in a consistent style (*human write in consistent style*). These responses expressed reviewers' belief that AI cannot replicate the natural flow and logical progression achieved by human writers through careful and critical thinking and appropriate sentence transitions.

> *"The discussion of the research method feels somewhat abrupt and lacks a smooth connection with the preceding and subsequent content." — Reviewer 6*

Conversely, 7% of responses indicated that AI produces well-structured sentences (*AI structure sentences better than human*). Responses also noted that AI often *uses conclusive statements at the end* (2%), and *follows an exact template* (1%), and frequently *uses transitional/bridged clauses in sentences* (1%). In contrast, 3% of responses mentioned that inexperienced human researchers often make mistakes and fail to produce well-structured sentences (*Inexperienced human writing mistakes*).

> *"The sentences are too well-structured to be human-written. It feels like this follows an exact writing template." — Reviewer 17*

Another interesting contradiction emerged regarding sentence length. While 13% of responses indicated that snippets involve AI tended to have *convoluted and long sentences*, 3% held the opposing view and attributed *convoluted and long sentences* to human writers. Additionally, one (1%) response expressed the reviewer's concern about *AI making grammar mistakes*, whereas 3% of responses indicated that snippets with grammar mistakes are more likely to be human-written (*human makes grammar mistakes*).

#### 4.4.2. *Theme 2: Word choice*

Another significant factor influencing reviewers' perceptions was the presence of marker phrases and words in both human- and AI-produced

**Table 4**
Multivariate Cumulative Linked Mixed Model analyses of factors impacting participants' perceived AI Involvement (0-completely human to 10-completely AI), with a random intercept per reviewer. Ordinal data are treated as is.

**DV = Perceived AI Involvement**

| Predictor | Estimates | Std. Error | z | *P* | OR (95%CI) | AIC |
|---|---|---|---|---|---|---|
| ***Content type*** | | | | | | 484.07 |
| Original | Reference | | | | | |
| Paraphrased | −1.002 | 0.432 | −2.319 | 0.016* | 0.367 (0.157, 0.856) | |
| Generated | 0.403 | 0.432 | 0.935 | 0.35 | 1.496 (0.642, 3.487) | |
| ***AI familiarity**** | | | | | | |
| Never | Reference | | | | | |
| Rarely | −1.297 | 0.658 | −1.973 | 0.049* | 0.273 (0.075, 0.992) | |
| Sometimes | −0.225 | 0.400 | −0.562 | 0.574 | 0.799 (0.365, 1.749) | |
| ***Peer-review experience*** | | | | | | |
| | 0 | 0.00127687 | −0.190 | 0.849 | 1.000 (0.997, 1.002) | |
| ***Disciplinary expertise*** | | | | | | |
| | 0.085 | 0.0625 | 1.365 | 0.172 | 1 (0.964, 1.231) | |

**DV = Perceived Honesty**

| Predictor | Estimates | Std. Error | z | *P* | OR (95%CI) | AIC |
|---|---|---|---|---|---|---|
| ***Disciplinary expertise*** | | | | | | 263.94 |
| | 0.163 | 0.0794 | 2.048 | 0.041* | 1.177 (1.007, 1.375) | |
| ***Content type*** | | | | | | |
| Original | Reference | | | | | |
| Paraphrased | 1.051 | 0.4701 | 2.233 | 0.026* | 2.861 (1.137, 7.19) | |
| Generated | 0.207 | 0.475 | 0.437 | 0.662 | 1.23 (0.485, 3.121) | |
| ***AI familiarity**** | | | | | | |
| Never | Reference | | | | | |
| Rarely | −0.151 | 0.931 | −0.163 | 0.871 | 0.86 (0.139, 5.327) | |
| Sometimes | 0.98 | 0.618 | 1.585 | 0.113 | 2.664 (0.793, 8.943) | |
| ***Peer-review experience*** | | | | | | |
| | 0 | 0.002 | 0.080 | 0.936 | 1 (0.996, 1.004) | |

**DV = Perceived Clarity**

| Predictor | Estimates | Std. Error | z | *P* | OR (95%CI) | AIC |
|---|---|---|---|---|---|---|
| ***Disciplinary expertise*** | | | | | | 288.05 |
| | 0.19 | 0.069 | 2.760 | 0.006** | 1.209 (1.057, 1.384) | |
| ***AI familiarity**** | | | | | | |
| Never | Reference | | | | | |
| Rarely | 0.71 | 0.771 | 0.921 | 0.357 | 2.034 (0.449, 9.213) | |
| Sometimes | 1.093 | 0.522 | 2.092 | 0.036* | 2.983 (1.072, 8.305) | |
| ***Peer-review experience*** | | | | | | |
| | −0.003 | 0.002 | −1.524 | 0.127 | 0.997 (0.994, 1.001) | |
| ***Content type*** | | | | | | |
| Original | Reference | | | | | |
| Paraphrased | 0.391 | 0.443 | 0.882 | 0.378 | 1.478 (0.62, 3.521) | |
| Generated | 0.319 | 0.456 | 0.701 | 0.483 | 1.376 (0.563, 3.362) | |

**DV = Perceived Compellingness**

| Predictor | Estimates | Std. Error | z | *P* | OR (95%CI) | AIC |
|---|---|---|---|---|---|---|
| ***AI familiarity**** | | | | | | 278.08 |
| Never | Reference | | | | | |
| Rarely | 1.793 | 0.681 | 2.634 | 0.008** | 6.007 (1.582, 22.821) | |
| Sometimes | 1.161 | 0.440 | 2.639 | 0.008** | 3.193 (1.348, 7.57) | |
| ***Disciplinary expertise*** | | | | | | |
| | 0.09 | 0.063 | 1.417 | 0.157 | 1.094 (0.966, 1.239) | |
| ***Peer-review experience*** | | | | | | |
| | 0.001 | 0.001 | 0.872 | 0.383 | 1.001 (0.998, 1.004) | |
| ***Content type*** | | | | | | |
| Original | Reference | | | | | |
| Paraphrased | 0.356 | 0.452 | 0.786 | 0.432 | 1.428 (0.588, 3.464) | |
| Generated | 0.322 | 0.460 | 0.699 | 0.485 | 1.38 (0.56, 3.402) | |

Note. *Only options selected by reviewers were displayed. Significance are displayed as follows: ***P < .001, **P < .01, *P < .05. DV = Dependent Variable. OR=Odds Ratio. CI=Confidence Interval. Predictors are arranged based on their contribution to the model, determined as global AIC. Predictors were ranked with the highest contribution (lowest AIC) appearing first. The Reference categories were selected to enhance result interpretability. For OR, a value greater than 1 indicates a positive relationship, and a value less than 1 indicates a negative relationship.

snippets. For instance, 26% of responses identified specific words and phrases (*AI uses marker words/phrases*) commonly used by AI, such as transition phrases like *"However, …"* and sentence structures like *" …, do-ing …"* In addition, terms such as *"leverage"* or *"state-of-the-arts"* were seen as indicators of AI writing due to their less common usage compared to simpler alternatives. Interestingly, reviewers' perceptions of these markers were not always consistent. While 3% of responses noted that contractions, parentheses for explanations, and colons to introduce multiple concepts were unique to human-written snippet (*human uses unique marker phrases/words*), these markers were also mentioned in other responses as the indication of AI-generated snippets. We include a full list of marker words mentioned by reviewersin

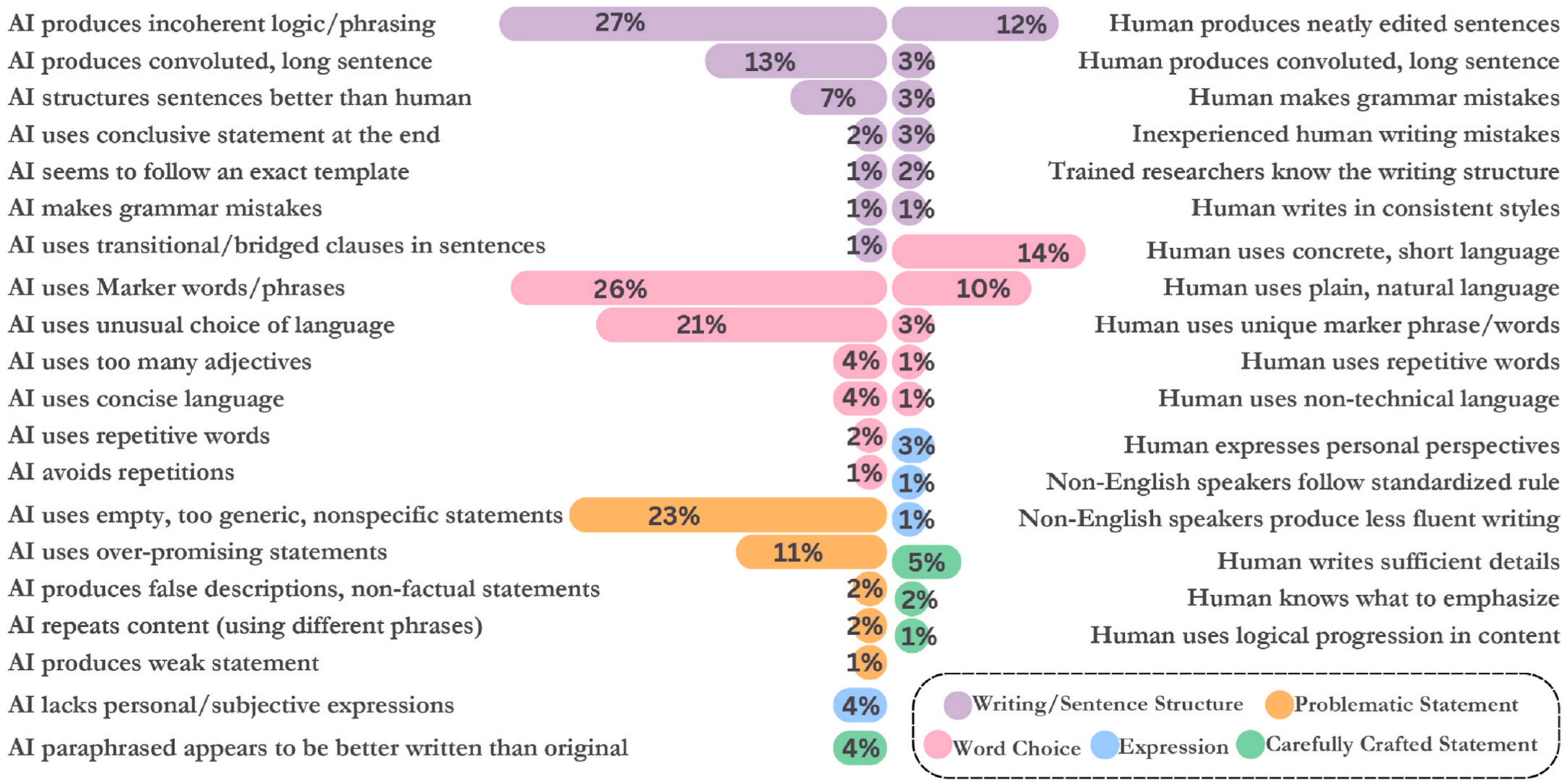

**Fig. 4.** Thematic Analysis Codebook. Synthesized reviewers' responses to the open-ended question (Q9): "What specifically in the snippet led you to believe it was written by human researcher(s) or generated by AI?". Each of the 17 reviewersanswered this question six times, resulting in 102 responses, with some responses mentioning multiple themes. Percentages calculated from the total responses (N = 102), with colors representing thematic codes in different themes (see bottom right). (For interpretation of the references to color in this figure legend, the reader is referred to the Web version of this article.)

Appendix section B.

> *"AI tends to construct sentences that often have a 'do-ing' in the second half." — Reviewer 9*

Beyond the identified marker words, reviewers also commented on broader language usage. Twenty-one percent of responses noted that unusual language choices, which made the sentences sound awkward or unnatural, were often perceived as a result of AI (*AI uses unusual language*). In addition, a small portion of responses (4%) criticized AI for relying too heavily on adjectives and resulting in an overly descriptive writing style. Conversely, some responses (10%) associated plain and natural language with human authors (*Human uses plain, natural language*). One response (1%) pointed out that human writing tends to incorporate more non-technical language compared to AI (*human uses non-technical language*) to cater to a broader audience.

> *"There are keywords: 'envision a future', 'excellent' (not good, not better, but excellent), 'high-fidelity' as an adjective for devices, 'exciting potential' (not only potential, but exciting one)." — Reviewer 14*

A small group of responses (4%) noted that AI tends to use concise language (*AI uses concise language*), while others (14%) associated snippets with concrete and succinct language with human authors (*human uses concrete, short language*). Additionally, some responses discussed the issue of repetitive wording: two (2%) mentioned it as a sign of AI-generated snippet (*AI uses repetitive words*), whereas one (1%) noted AI actively avoids repetitions (*AI avoids repetitions*), and another (1%) associated repetitive words with human-written snippets (*human uses repetitive words*).

#### 4.4.3. Theme 3: Problematic Statement

Reviewers raised various concerns regarding statements in the snippets. The most common issue was the presence of *generic and non-specific* statements, which were perceived as being associated with AI in 23% of responses. Additionally, 11% of responses noted *over-promising statements* as snippets that resulted from AI. Concerns about factual accuracy were also raised, with two responses (2%) noted snippets with *false descriptions and non-factual statements* as produced by AI.

> *"It is very generic and does not give concrete examples of what the authors do in the paper." — Reviewer 4*

Interestingly, two responses (2%) believed that AI often repeats content with different phrasing that merely summarizes earlier paragraphs without further elaboration. One response (1%) noted AI often produce weak statements (AI *produces weak statement*) that lack supporting evidence or being poorly developed.

#### 4.4.4. Theme 4: Expression

Reviewers assessed how well the snippets conveyed human emotions, opinions, and subjective experiences. Three percent of responses indicated that human authors use evocative words and figurative language to convey personal perspectives (*Human expresses personal perspectives and understanding in writing*), and 4% of responses associated snippets lacking personal and subjective expressions with AI (*lacks personal expressions*).

> *"There is a humble and stumble feel to the writing, which makes it feel like human." — Reviewer 17*

In addition, one response (1%) linked snippets strictly following the standardized grammar rules and sentence structure to non-English-speaking writers (*human non-English speakers follow standardized rules*). On the contrary, another response (1%) noted that non-native speakers might produce less fluent writing (*human non-English speakers could produce less fluent writing*). These contradictory perceptions can lead to inaccurate conclusions about AI involvement.

#### 4.4.5. Theme 5: Carefully Crafted Statement

Interestingly, 4% of responses admired the expertise in the writing of some snippets and perceived these snippets as *"the work of experienced researchers"* (*AI paraphrased content appears to be written by an experienced researcher*). However, these snippets were actually paraphrased using GenAI.

> *"I do think it was written by a human with good language skills."* — *Reviewer 14*

In addition, 5% of responses highlighted that human writing typically incorporates sufficient details and evidence to support statements (*human writes sufficient details*). Two responses (2%) noted that human authors emphasize key research points through strategic sentence structure, word choice, and transitional phrases, rather than presenting redundant information before reaching the main points. Similarly, one response (1%) expressed the belief that human authors are more likely to use logical progression (*Human uses logical progression in content*), and carefully present information in a smooth, clear, and coherent structure from research motivation and design to findings and discussions.

## 5. Discussion

Our study supports the concern raised by Tu et al., (2024) and extends prior research on the difficulty humans have in distinguishing between AI- and human-authored content in general contexts like news, jokes, and health information (e.g., Gao et al., 2023; Irene, 2024; Ragot et al., 2020). We found that such an inability also applies to peer reviewers of research publications. Our qualitative analysis highlighted contradictory perceptions among reviewers, where some reviewers identified lengthy sentences, concise language, repetition, and standardized grammar as indicators of AI authorship, while others perceived these as signs of human writing. However, despite these conflicting views, reviewers' judgments of the manuscript and the presented research remained consistent.

In fact, unlike prior research that identified people's tendencies toward AI aversion or appreciation (Christer, 2017; Graefe et al., 2018; Irene, 2024; Longoni et al., 2022), our study found that academia and industry professionals did not exhibit clear negative or positive opinions about the manuscript and its presented research across the three types of snippets, despite varying perceptions of AI involvement. This finding indicates that assessments of research extend beyond writing quality alone. While clarity, conciseness, and coherence are important, other factors such as novelty, methodological transparency, result validity, and contribution to the field also significantly influence reviewers' judgments (The 2023 ACM CHI Conference, 2023). Thus, our results suggest that when these aspects are well-addressed, the use of GenAI in writing does not necessarily bias reviewers' evaluations.

Furthermore, our regression analysis indicated that reviewers perceived less AI involvement and higher honesty in AI-paraphrased snippets. Reviewers with greater disciplinary expertise and AI familiarity rated higher levels of honesty, clarity, and compellingness across all snippet types. This result contrasts with previous studies on non-research writing contexts, where experts found algorithmic advice less trustworthy (van der Kaa & Krahmer, 2014) and those familiar with the algorithm were less receptive to its suggestions (Logg et al., 2019). Our qualitative results further showed that reviewers appreciated GenAI's ability to produce well-structured and clear snippets. This perception suggests that GenAI can be valuable for enhancing the presentation of their research through writing. However, reviewers often found issues of lacking logical progression, supporting evidence for statements, and emphasis on key research points in snippets resulted from AI. These issues highlight the limitations of GenAI in areas requiring critical thinking, logical reasoning, and nuanced understanding of the research field. Conversely, reviewers noted that human researchers are good at providing detailed evidence and explanation, and strategically emphasizing key points within the manuscript's logical flow. Given that increased human involvement in AI-produced content fosters greater ownership and responsibility (Fiona et al., 2024; Pierce, Kostova, & Dirks, 2003), we thus recommend a human-in-the-loop approach to AI-assisted writing to ensure logical, clear, and accurate research manuscripts.

Overall, our study suggests that while AI can be valuable in enhancing research communication by improving structure and clarity of its presentation, human researchers' oversight remains crucial to ensure a well-structured, logically sound, and informative final manuscript.

### 5.1. *The discussion of the research method implications for researchers who submit to peer-reviewed venues*

Through the perspective of top-tier HCI conference peer-reviewers, our quantitative and qualitative analyses revealed themes that alleviate researchers' concerns about disclosing AI use in manuscript submissions. From reviewers' responses, we identify insights on the appropriate ways to augment research writing with GenAI, and demonstrate that responsible and transparent use of GenAI can enhance the quality of research presentation in writing without damaging reviewers' perceptions on the underlying research. Our reviewers agreed on GenAI's ability to produce well-structured and readable sentences, which highlights its potential benefits for novice researchers and non-native English speakers who struggle with writing. GenAI can act as an assistant to improve the overall grammar, sentence structure, and clarity of their manuscripts. However, researchers should not overly rely on GenAI, as our reviewers pointed out its limitations, such as a lack of logical flow, insufficient supporting evidence, and the use of inaccurate or non-factual statements—a fundamental problem in the underlying language generation models (Achiam et al., 2023; Ji et al., 2023). This issue can be particularly harmful to those less familiar with the research domain or with limited English proficiency.

Literature has identified researchers' concerns that AI cannot uncover nuanced insights from data and can lead to generic themes that overlook data complexity and diversity (Jie, 2024). In the context of research writing, our reviewers echoed this sentiment and noted that AI often replicates content with various generic statements and lack rele-vant details to the research. This result suggests that current GenAI cannot independently perform meaningful and comprehensive data in-terpretations and therefore should not replace the critical thinking and in-depth analysis human researchers bring into the writing. Beyond this, reviewers valued the emotions and subjective expressions conveyed by human authors, and appreciated the *"human touch"* in research writing. This echoes the finding from Clerwall (Christer, 2017) on news articles. While our reviewers found snippets resulted from AI were easier to read, they also noted a sense of monotony due to the repetitive and stan-dardized structure and style. The personal and subjective elements from human researchers make reviewers see academia as a diverse, curious, and collaborative community, rather than a collection of impersonal paper-producing machines. This finding further reinforces the impor-tance for human researchers to act as the primary driver of the writing process even with AI assistance.

In summary, our findings show that using GenAI for writing augmentation does not negatively impact reviewers' perceptions. Based on our findings, we strongly advocate for a balanced approach to GenAI use in academic writing. Researchers should make use of GenAI as a tool for enhancing readability and reorganizing research knowledge. How-ever, they should remain in their role as the primary intellectual drivers of their work. We emphatically recommend that researchers: (1) Openly disclose their use of GenAI in manuscript preparation to foster trans-parency and trust in the academic community. (2) Carefully review and fact-check all AI-augmented content, so that the facts are correct and the output is aligned with their intended arguments. (3) Preserve the *"human touch"* in their writing, which our study shows resonates strongly with reviewers and keeps the collaborative spirit of academic discourse. (4) Use GenAI judiciously to enhance—not to replace—their critical thinking and unique insights. These guidelines enable re-searchers to create clearly presented research while mitigating the risk of false or generic GenAI statements. This approach maintains research integrity and aligns with evolving ethical standards in academic publishing (Panel on Responsible Conduct of Research, 2021). Responsible and transparent use of GenAI will be crucial to preserve the quality and credibility of peer-reviewed research.

### 5.2. Implications for peer-reviewers who review research manuscripts

While research venues permit the use of GenAI as writing assistants, these tools must be accompanied by human author oversight and verification (Elsevier.). As demonstrated in Fig. 4, our study revealed that reviewers identified similar issues in both AI- and human-written snippets, such as redundant sentences, overly generic statements, and marker phrases (see Appendix section B). Thus, these problems are common in both human-written and AI-augmented manuscripts and cannot be used as reliable evidence of AI involvement. Despite the availability of algorithm-based AI detectors, literature shows these tools often penalize individuals with limited linguistic proficiency (Liang et al., 2023), which directly contradicts our reviewers' perception that snippets resulted from AI use *"flowery"* language. This contradiction highlights that neither existing AI-detectors nor reviewers' personal strategies are reliable in detecting GenAI. Given our findings, we strongly advise reviewers to refrain from speculating about GenAI involvement in manuscripts, because both human intuition and AI detectors have proven unreliable in this regard. Instead, we recommend that reviewers: (1) focus exclusively on the manuscript's scientific merit (e.g., validity of methods, robustness of results, significance of contributions), (2) evaluate the manuscript's coherence, clarity, and effective communication of research findings regardless of perceived authorship method, (3) base their assessment on the strength of arguments and quality of evidence presented (not language use or writing style assumptions), and (4) if concerns about academic integrity arise, to address these through established channels.

Although our reviewers did not show clear positive or negative perceptions across the three snippet types, their perceptions may become more diverse as GenAI functionalities continue to proliferate, its use in research activities continues to grow, and its counter-movements (e.g., PauseAI[10]) continue to rise in influence. Literature indicates that acceptance rates of manuscripts from non-English-speaking countries are significantly lower than those from English-speaking countries (Ehara & Takahashi, 2007). Thus, it is understandable that non-English-speaking researchers might use GenAI to ensure their manuscripts conform to standard scientific English, are clear, and appealing to reviewers, and can compete with those from native English-speakers. While reviewing a manuscript entails the responsibility of assessing and ensuring the quality of published research (Elsevier, n.d.-b), we emphasize that the fundamental principles of peer review remain unchanged—even when GenAI is used in academic writing. Reviewers should reaffirm their commitment to the collaborative nature of a peer review, which aims to guide researchers toward excellence rather than merely critiquing their work (Mohan, 2006). It is imperative that reviewers remain objective on the manuscript's scientific merit, methodological rigour, and contributions to the field. They should always provide constructive feedback that enhances the quality of the work and supports an author's development. Reviewers will have to recognize that GenAI use may be an assistive tool for non-native speakers to help them overcome an existing language handicap. Reviews should be adapted to acknowledge the evolving nature of academic writing, where the lines between human and AI-assisted content are becoming increasingly blurred.

### 5.3. Future enforcement of ethical use of GenAI in research writing

Our study sheds light on the complexities of regulating and enforcing ethical GenAI use in research writing. Our findings revealed the unreliability of strategies that human reviewers use to distinguish between AI and human authorship. Together with the unreliable result from existing GPT detectors Liang et al. (Liang et al., 2023), we highlight that current human and algorithm-based methods for identifying AI produced content can increase biases and inequities in academic publishing. We argue that AI-detecting tools, in their current state, should be used cautiously and only as supplementary information, not as definitive evidence of AI involvement in manuscript writing. The primary focus should remain on human reviewers' critical assessments of research quality and contribution. Concurrently, we recommend that academic institutions and publishing venues invest in educating reviewers about the capabilities and limitations of GenAI, as well as the potential biases in both human and algorithmic detection methods. This education should emphasize the importance of evaluating manuscripts based on their scientific merit, regardless of perceived AI involvement. A more nuanced understanding of GenAI among reviewers promotes fairer evaluations of research manuscripts and maintains the integrity of the peer-review process in the continuously evolving GenAI space.

Several reviewers expressed their concern in the end-survey comments about the pressure in academia to produce numerous papers quickly for job security and career progression. This demand leads researchers to prioritize short, impactful studies over longitudinal work. GenAI exacerbates this issue by speeding up the writing process, which can undermine careful and thoughtful research and writing. This comment echoes the sentiment from literature that AI will create a negative feedback loop for researchers who write manually and lead to a drought of human-created content (Irene, 2024; Max, 2024). Given these concerns, it is crucial to balance the advantages of AI-augmented writing with the preservation of human authorship values. We propose a multi-faceted approach that is both realistic and impactful. Academic venues need to update their submission guidelines. Voluntary disclosure of GenAI prompts without repercussions would make the process of AI use more transparent. However, human oversight and critical thinking should remain the most important components of the review process. To facilitate this, institutions and funding bodies should provide ethical guidelines for using GenAI in research. The mindset shift required for authors here would be to focus more on research quality, impact, and innovation instead of publication quantity. This shift would disincentivize abuse of GenAI and support longer-term, more comprehensive studies. In addition, better training for reviewers on what GenAI can and cannot do would also let them focus more on evaluating the research's value.

More research is needed on the long-term effects of GenAI writing and research quality to inform any future policy changes. However, full regulation and verification of GenAI may not be feasible or even desirable, but a research culture that values thorough, impactful work while acknowledging the role of new technology should be. Our goal as researchers should be to mitigate the potential negative effects of GenAI on research quality and human authorship while still benefiting from the capabilities of this new technology to enhance academic writing.

### 5.4. Limitations and opportunities for future research

Our study has limitations that offer several opportunities for future research. First, our primary limitation is its sample size. While our mixed-methods approach gave deep and rich insights, the small number of participants limits the precision of our quantitative estimates. This constraint reflects the challenges in recruiting professional peer reviewers (Henderson et al., 2020, pp. 957–959), a hurdle likely to persist in future studies. Still, our findings offer early and valuable insights into how reviewers see GenAI in academic writing. Future research could explore other approaches. For example, it could analyze acceptance and rejection patterns before and after GenAI adoption. This would add to our findings, even though such data might be equally difficult to obtain. Second, our study focused on abstracts, not full papers. This approach let us examine varied AI snippets across research areas while maintaining survey manageability for professional reviewers. However, it may not fully capture reviewers' judgments of complete manuscripts. Future research should extend this investigation to full papers or more extensive snippets. This could reveal more nuanced perceptions of AI-augmented academic writing. Third, our sample's limited familiarity with AI in writing likely reflects the current reviewer population, given the ongoing controversies surrounding GenAI use in academia. As GenAI becomes more commonplace in research activities, future studies may reveal evolving perceptions among reviewers. This presents an opportunity for longitudinal research to track changes in reviewer attitudes and practices over time. Fourth, our study also suffer from common limitations of empirical research. Although we instructed reviewers not to look up the full papers in literature databases, we cannot entirely prevent this. Additionally, our data relies on self-reporting, which is subject to the reviewers' honesty and self-awareness. Our study may be subject to social desirability bias. Reviewers are potentially underreporting their own GenAI use because of perceived stigma. However, we expect that our findings will help normalize discussions about GenAI in academic writing and, in turn, encourage more open disclosure in future research. Lastly, we studied general reviewer perceptions across disciplines, with 35% of reviewers specializing in games research, likely due to our research team's majority background in this field. Future research should explore how specific disciplines view GenAI use. This is especially true in disciplines with writing styles that could be misidentified as AI-generated. It would provide a more nuanced understanding of GenAI's impact on peer review processes.

[10] PauseAI Proposal. https://pauseai.info/proposal.

## 6. Conclusion

Our paper presents a snippet-based online survey examining reviewers' perceptions of human-written, AI-paraphrased, and AI-generated snippets. We surveyed 17 experienced peer-reviewers from top-tier HCI conferences and found their struggle in distinguishing between AI-processed and human-written snippets but their judgments on the manuscript and the underlying research did not significantly vary. Our results indicate that responsible and transparent use of GenAI can enhance research presentation quality without negatively impacting reviewers' perceptions. Given the current unreliability of AI detection by reviewers and AI-detection tools, we advocate for reviewer guidelines that promote impartial evaluations of submissions, regardless of any personal biases towards GenAI. Our findings encourage researchers to transparently disclose their AI use in manuscripts without the fear of damaging reviewers' perception. Based on our findings, we that researchers must maintain their authorship and control over the writing process, even when using GenAI assistance.

## CRediT authorship contribution statement

**Hilda Hadan:** Writing – review & editing, Writing – original draft, Visualization, Validation, Software, Resources, Project administration, Methodology, Investigation, Formal analysis, Data curation, Conceptualization. **Derrick M. Wang:** Software, Formal analysis. **Reza Hadi Mogavi:** Writing – review & editing, Data curation, Conceptualization. **Joseph Tu:** Data curation. **Leah Zhang-Kennedy:** Writing – review & editing, Supervision, Funding acquisition. **Lennart E. Nacke:** Writing – review & editing, Supervision, Funding acquisition, Data curation, Conceptualization.

## Declaration of competing interest

The authors declare the following financial interests/personal relationships which may be considered as potential competing interests: Dr. Lennart Nacke reports financial support was provided by Natural Sciences and Engineering Research Council of Canada. Dr. Leah Zhang-Kennedy reports financial support was provided by Natural Sciences and Engineering Research Council of Canada. Dr. Lennart Nacke reports financial support was provided by Social Sciences and Humanities Research Council of Canada. Dr. Lennart Nacke reports financial support was provided by Canada Foundation for Innovation. If there are other authors, they declare that they have no known competing financial interests or personal relationships that could have appeared to influence the work reported in this paper.

## Acknowledgements

We thank our participants for taking a part in our study and sharing their insightful thoughts and opinions. We acknowledge that we used Grammarly's AI assistant, Claude 3.5 Sonnet, and Hemingway's AI Editor for spelling, grammar, punctuation, and clarity editing. Google Gemini was used to process the snippets as our study materials, and to paraphrase our initial abstract draft using example prompts from "AI-paraphrased snippets" to enhance its clarity and readability. Our decision to use Gemini on our abstract was informed by our findings. We also intend to inspect reviewers' reactions on our GenAI-paraphrased abstract during an actual peer-review process. Our manuscript was fully verified and edited by our research team. Our research team takes full responsibility for the content of the publication. We did not use generative AI for data collection, analysis, or image generation. Figures in this manuscript were created using Python seaborn package and pre-built templates on Canva, and statistical analyses were conducted using R.

This research was supported by the Natural Sciences and Engineering Research Council of Canada (NSERC) Discovery Grant (#RGPIN-2022-03353 and #RGPIN-2023-03705), the Social Sciences and Humanities Research Council of Canada (SSHRC) Insight Grant (#435-2022-0476), the Canada Foundation for Innovation (CFI) JELF Grant (#41844). Any opinions, findings, and conclusions or recommendations expressed in this material are those of the author(s) and do not necessarily reflect the views of the NSERC, the CFI, nor the University of Waterloo.

## Appendix A. Supplementary Material

Supplementary material to this article can be found online at https://doi.org/10.1016/j.chbah.2024.100095.